\documentclass[a4paper,twoside]{article}

\usepackage{epsfig}
\usepackage{subcaption}
\usepackage{calc}
\usepackage{amssymb}
\usepackage{amstext}
\usepackage{amsmath}
\usepackage{amsthm}
\usepackage{multicol}
\usepackage{pslatex}
\usepackage{apalike}
\usepackage{graphicx}
\usepackage{makecell}
\usepackage{multirow}
\usepackage{diagbox}
\usepackage{amsfonts}
\usepackage{bbding}
\usepackage{verbatim}
\usepackage{amsmath}
\usepackage{bm}
\usepackage{url}
\usepackage{SCITEPRESS}     
\usepackage{float}
\usepackage[table]{xcolor}
\newcommand{\tabincell}[2]{\begin{tabular}{@{}#1@{}}#2\end{tabular}}

\begin{document}

\title{TVNet: Temporal Voting Network for Action Localization}

\author{\authorname{Hanyuan Wang\orcidAuthor{0000-0002-9349-9597}, Dima Damen\orcidAuthor{0000-0001-8804-6238}, Majid Mirmehdi\orcidAuthor{0000-0002-6478-1403} and Toby Perrett\orcidAuthor{0000-0002-1676-3729}}
\affiliation{Department of Computer Science, University of Bristol, Bristol, U.K.}
\email{\{hanyuan.wang, dima.damen, toby.perrett\}@bristol.ac.uk, majid@cs.bris.ac.uk}
}


\abstract{We propose a Temporal Voting Network (TVNet) for action localization in untrimmed videos. This incorporates a novel Voting Evidence Module to locate temporal boundaries, more accurately, where temporal contextual evidence is accumulated to predict frame-level probabilities of start and end action boundaries. Our action-independent evidence module is incorporated within a pipeline to calculate confidence scores and action classes. We achieve an average mAP of 34.6\% on ActivityNet-1.3, particularly outperforming previous methods with the highest IoU of 0.95. TVNet also achieves mAP of 56.0\% when combined with PGCN and 59.1\% with MUSES at 0.5 IoU on THUMOS14 and outperforms prior work at all thresholds. Our code is available at \url{https://github.com/hanielwang/TVNet}.}

\onecolumn \maketitle \normalsize \setcounter{footnote}{0} \vfill


\section{\uppercase{Introduction}}
\label{sec:introduction}
While humans are capable of identifying event boundaries in long videos \cite{humanevent}, 
in a remarkably consistent fashion, current approaches fall short of optimal performance~\cite{2017SSN,ReFRCNN,Lin_2018_ECCV,PGCN,Lin_2019_ICCV,GTAD,FastLearning,BSN++} despite being trained on large and varied datasets \cite{THUMOS14,Activitynet}, primarily due to varying action durations and densities \cite{Lin_2018_ECCV,MGG,BSN++}.

Current state-of-the-art action-recognition methods attach equal importance to each frame when determining where action boundaries occur \cite{Lin_2018_ECCV,Lin_2019_ICCV,MGG,BSN++}. Intuitively, a frame near the start of an action should be better suited to predicting the start point of the action than a frame in the middle of the action, and similarly for end points. One would expect this to result in more accurate boundary locations. In this paper, we incorporate this intuition, while still utilising all frames in the untrimmed video, such that each and every frame can contribute evidence when predicting boundary locations. Distinct from previous approaches, we weight this evidence depending on its distance to the boundary. We propose the Temporal Voting Network (TVNet), which models  relative boundary locations via contextual evidence voting. Specifically, this contains a proposed Voting Evidence Module to locate temporal boundaries based on accumulating temporal contextual evidence.

Our key contributions can be summarized as follows: (1) We introduce a novel voting network, TVNet, for accurate temporal action localization.
(2) We evaluate our method on two popular action localization benchmarks: ActivityNet-1.3 and THUMOS14. Our method achieves significant improvements, especially at high IoU thresholds, demonstrating more precise action boundaries.
(3) We perform a detailed ablation, finding that both temporal context voting and attention learning are crucial to TVNet's performance.

\section{\uppercase{Related Work}}

Methods for action recognition assume trimmed videos as input, which they directly classify \cite{Two-stream,2015Learning,2017Quo,SlowFast}.  
Instead, temporal action localization works aim to locate actions in untrimmed videos as well as classify them.
Most works, like ours, investigate proposal generation \emph{and} proposal evaluation \cite{2017SSN,ReFRCNN,Lin_2018_ECCV,Lin_2019_ICCV,MGG,2019Gaussian,GTAD,BC-GNN,Uty,BSN++}, but some just focus on proposal evaluation, such as~\cite{PGCN,MUSES}. 

Notable studies include MGG \cite{MGG}, which uses  frame-level action probabilities to generate segment proposals. BSN \cite{Lin_2018_ECCV} and BMN \cite{Lin_2019_ICCV} generate proposals based on boundary probabilities, using frame-level positive and negative labels as supervision. However, their probabilities are calculated independently and each temporal location is considered as an isolated instance, leading to sensitivity to noise.  
Therefore, several works focus on exploiting rich contexts. G-TAD \cite{GTAD} proposes a model based on a graph convolutional network to incorporate multi-level semantic context into video features. GTAN \cite{2019Gaussian} involves contextual information with the feature using Gaussian kernels. The recently introduced BSN++ \cite{BSN++} uses a complementary boundary generator to extract both local and global context. However, these works ignore that different contextual frames have different degrees of importance, which may lead to insufficient exploitation of context, and generate imprecise boundaries, especially in complex scenes. 

In contrast, a context-aware loss function for action spotting in sports videos is proposed in \cite{Soccer-Spotting}, which treats frames according to their different temporal distances to the ground-truth moments. This method is designed to detect an event has occurred, and does not perform as well as specialist action localization methods when adapted to predict precise starting and ending boundaries. 
Our work is inspired by the notion of voting to incorporate contextual information, which was used for the task of moment localization in action completion~\cite{Farnoosh2018,Farnoosh2019}. However, these works aim to recognise a single moment for retrieval, rather than start and end times of actions. We thus offer the first attempt to incorporate voting into action localization. We detail our method next.

\section{\uppercase{Method}}

\begin{figure*}[t]
	\centering
	\includegraphics[height=0.28\textheight]{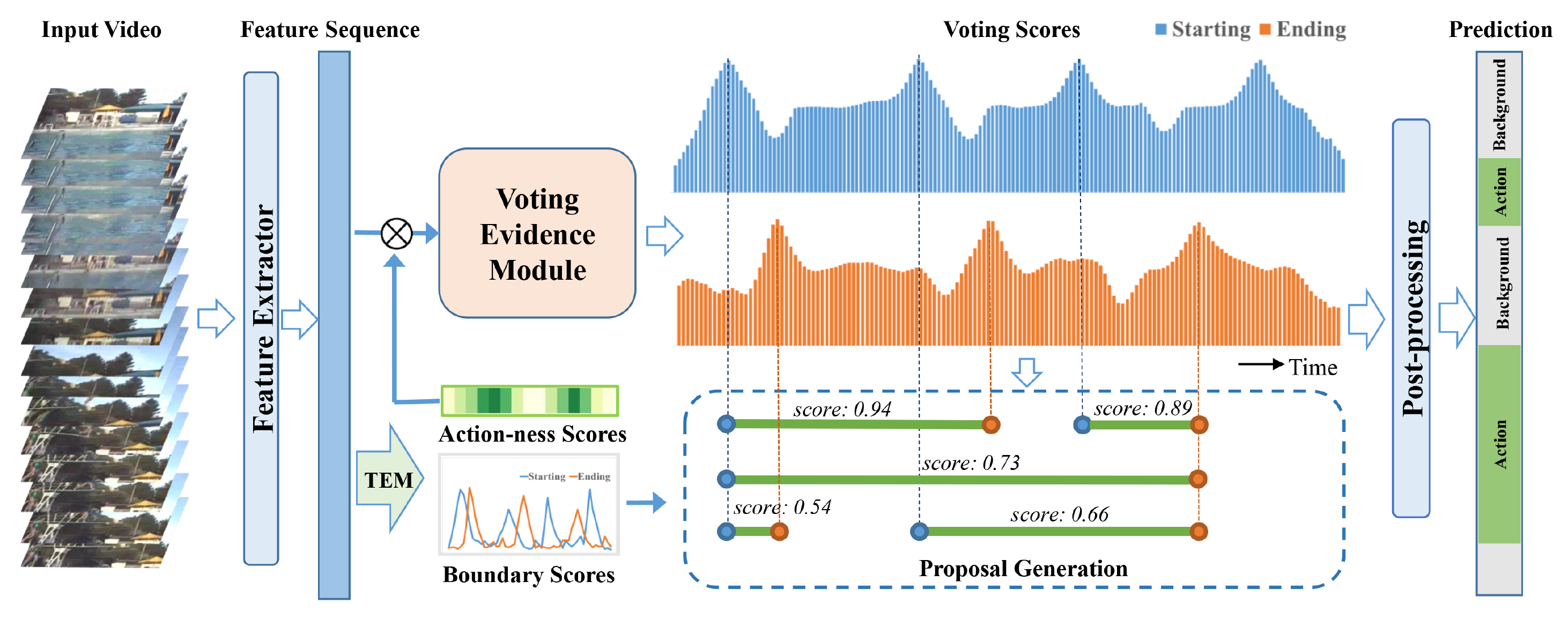}
    \caption{Overview of our proposed TVNet. Given an untrimmed video, frame-level features are extracted. Our main contribution, the Voting Evidence Module, takes in this feature sequence and outputs sequences of starting and ending voting scores. Local maxima in these voting scores are combined to form action proposals, which are then scored and classified. }  
	\label{fig:framework}
\end{figure*}

Our proposed method, TVNet, takes a feature sequence as input. It produces a set of candidate proposals using a Voting Evidence Module (VEM), where each frame in the sequence can contribute to boundary localization through voting, whether itself is a boundary frame or not.
We then calculate confidence scores and classify these candidate proposals with an action classifier.
An overview of TVNet is illustrated in Figure \ref{fig:framework}.

Section \ref{subsec:def} offers a formal problem formulation. Section \ref{subsec:voting evidence module} introduces our main contribution, the Voting Evidence Module. Finally, Section \ref{proposals generation and post-processing} discusses how the VEM is used within the full proposal generation process.

\subsection{Problem Definition}
\label{subsec:def}
Given an untrimmed video $V$ with length $L$, the feature sequence is denoted as $F =  \left\{f_{t}\right\}^{T}_{t=1}$ with length $T$, extracted at rate $\frac{L}{T}$. Annotations of action instances in video $V$ can be denoted as $\Psi= \{(s,\  e,\  a)\}^K_{k=1}$, where $K$ is the total number of ground truth action instances, and $s$, $e$ and $a$ are starting time, ending time and label of action instance, respectively. The goal of the temporal action localization task is to predict a set of possible action instances $\hat{\Phi}=\left\{(\hat{s},\ \hat{e},\ \hat{c},\ \hat{a}) \right\}^M_{m=1}$. Here, $\hat{s}$ and $\hat{e}$ are the starting and ending boundaries of the predictions, $\hat c$ is a confidence score for the proposal,  $\hat{a}$ is the predicted class, and $M$ is the number of predicted action instances. The annotation set $\Psi$ is used to calculate the loss for training. The predicted instance set $\hat{\Phi}$ is expected to cover $\Psi$ with high overlap and recall, so $M$ is likely to be larger than $K$.

\subsection{Voting Evidence Module (VEM)}
\label{subsec:voting evidence module}

Our premise is that each frame in the sequence offers information that would assist in locating the start and end of nearby frames. 
For each frame, we aim to predict its relative signed distance (in frames) to the start of an action of interest.
Assume a frame is 5 frames past the start of the action. We denote this relative distance `-5', indicating the start of the action is 5 frames ago.
In contrast, when a frame precedes the start of the action by say 2 frames, we denote `+2'.
Similarly, we can predict the relative signed distance from a frame to the end of an action.
To better understand our objective, we consider two cases. First, if a frame is within an action,
the optimal model would be able to predict the relative start frame and the end frame of the ongoing action, from this frame.
The second case is when the frame is part of the background.
The optimal model can still predict the elapsed time since the last action concluded as well as the remaining time until the next action starts. 

Evidently, a single frame's predictions can be noisy, so we utilise two techniques to manage the noise. The first is to only make these predictions within local temporal neighbourhoods, where evidence can be more reliable, and the second is to accumulate evidence from all frames in the sequence. We describe these next.

\begin{figure}[t]
\centering 
\centering
\includegraphics[height=0.25\textheight]{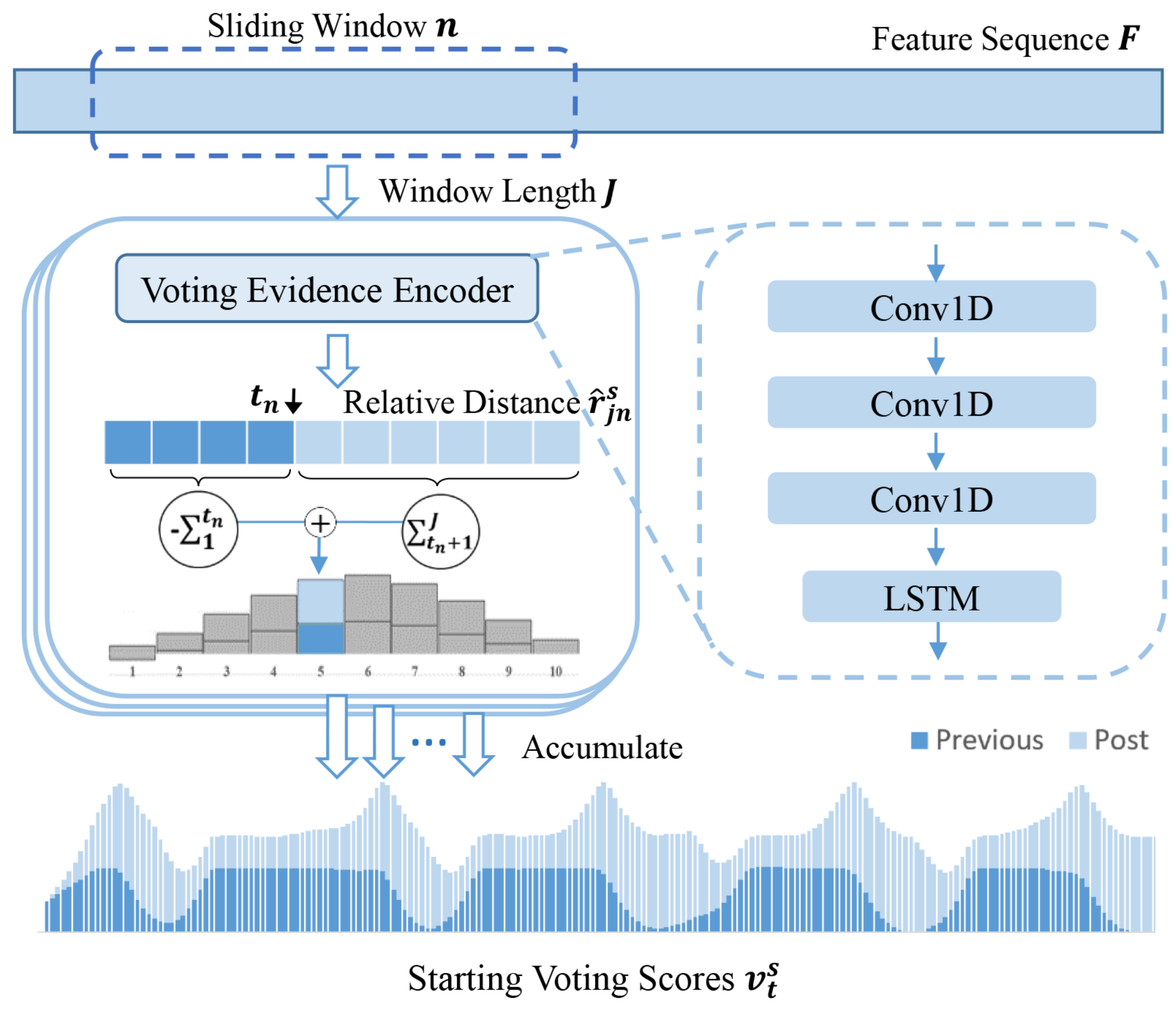}
\caption{Illustration of our proposed Voting Evidence Module. We accumulate evidence from all frames to calculate boundary scores for starting.}  
\label{fig:voting module}
\end{figure}

\noindent {\bf Voting Evidence Encoder.} Given the feature sequence $F$, we use a sliding window of length~$J$.
Accordingly, we only use neighbourhood of size $J$ in support of temporal boundaries. This is passed to one-dimensional temporal convolutional layers, in order to attend to local context, as well as an LSTM of past input for sequence prediction.
The network construction is illustrated in Figure \ref{fig:voting module}.
The output prediction at each frame would be $\hat R = \left\{(\hat{r}^{s}_{j}, \hat{r}_j^e)\right\}^{J}_{j=1}$, and it denotes the relative distance $\hat{r}$ to the \emph{closest} start/end to frame $j$.

We supervise the training of the VEM from the ground-truth. 
\begin{figure} [ht!]
\centering 
\centering
\includegraphics[height=0.245\textheight]{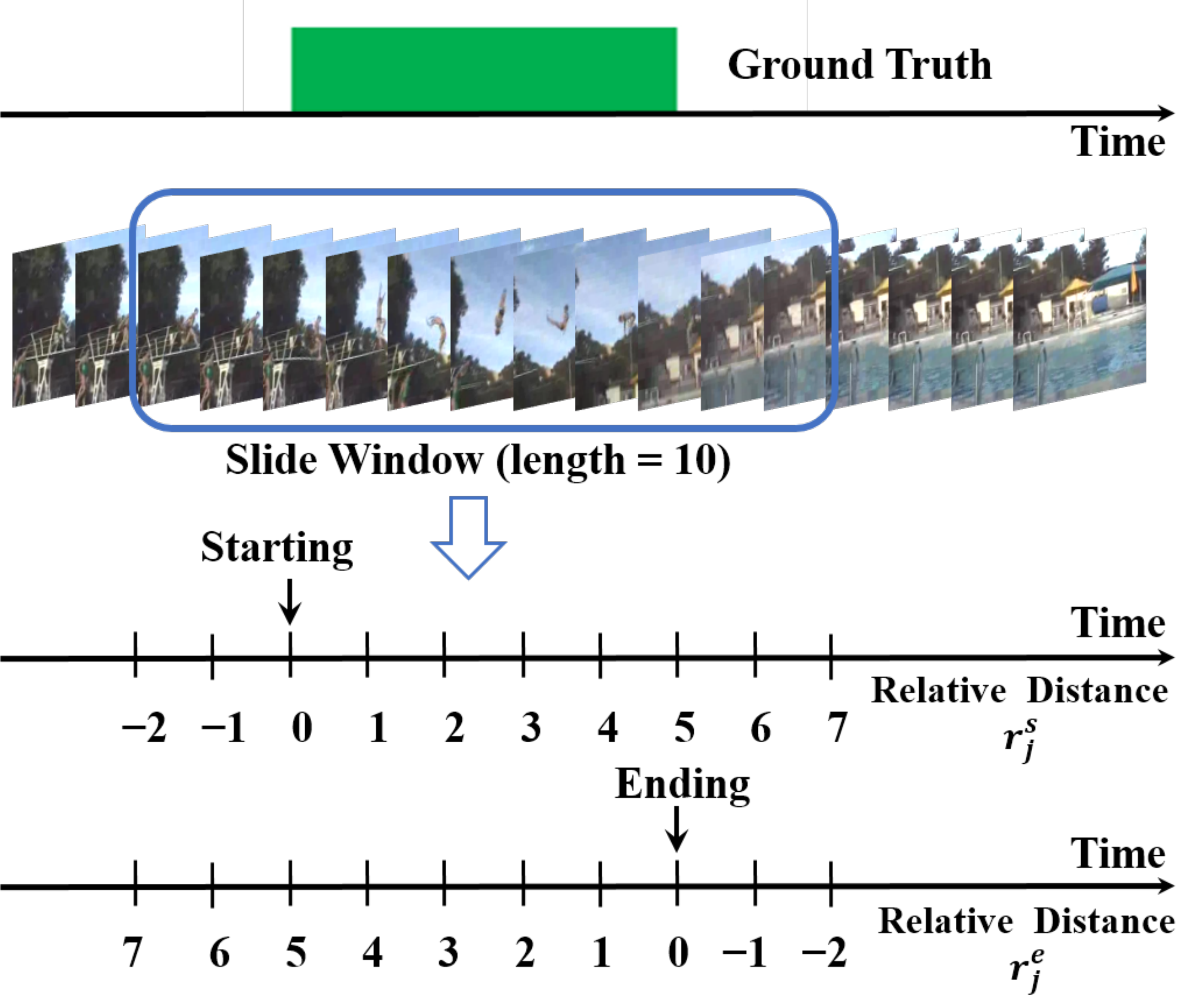}
\caption{Supervision for relative distances of starting and ending action boundaries within a sliding window, from ground truth.}
\label{fig:relative distance}  
\end{figure}
For each window, we use the relative distance between the current temporal location and the ground-truth boundaries as training labels.
As shown in Figure \ref{fig:relative distance}, we have a ground-truth relative distance set ${R = \left\{(r^{s}_{j},r^{e}_{j})\right\}^{J}_{j=1}}$. The values of $r^{s}_{j}$, $r^{e}_{j}$ reflect the distance between location $j$ and the closest start location $s^*$ as well as the closest end $e^*$. Therefore, the ground-truth relative distance is defined as $r^{s}_{j} =j - s^* $ for starting, and $r^{e}_{j} = e^* -j $ for ending.

We normalize the relative distance value to -1 to 1. We train the VEM for starting and ending separately, using the following MSE losses, shown here for a single window:
\begin{equation}
\begin{aligned}
{L}^{s}=\frac{1}{J}\sum_{j=1}^{J} (\hat{r}^{s}_{j}-r^{s}_{j})^{2} \qquad \mbox{and} \qquad 
{L}^{e}=\frac{1}{J}\sum_{j=1}^{J} (\hat{r}^{e}_{j}-r^{e}_{j})^{2}.
\end{aligned}
\end{equation} 

\noindent {\bf Voting accumulation.} We accumulate the predicted relative distance votes from all frames in the sequence as 

\begin{equation}
    \begin{aligned}
v^{s}_{t}&=&\sum_{n=1}^{N} \left( \sum_{j=1}^{t_n} (-\hat{r}^{s}_{jn})+\sum_{j=t_n+1}^{J} \hat{r}^{s}_{jn}\right) ~,\\
v^{e}_{t}&=&\sum_{n=1}^{N} \left( \sum_{j=1}^{t_n} \hat{r}^{e}_{jn}+\sum_{j=t_n+1}^{J} (-\hat{r}^{e}_{jn})\right) ~,
    \end{aligned}
\end{equation}
where $v^{s}_{t}$, $v^{e}_{t}$ are the voting scores for our predictions, $N$ is the number of windows sliding over location $t$ for which we are voting, $r^s_{jn}$ is the $j^{th}$ relative distance in the $n^{th}$ window, and similarly for $r^e_{jn}$, with $t_n$ being the corresponding location of frame $t$ in window $n$. 
The higher the voting score at location $t$, the more likely $t$ is to be a boundary location. The sequences of voting scores for starting and ending are denoted as $V^{s} =\left\{v^{s}_{t}\right\}^{T}_{t=1}$ and $V^{e} =\left\{v^{e}_{t}\right\}^{T}_{t=1}$.

From $V^{s}$, $V^{e}$, we can generate action proposals, explained next in Section \ref{proposals generation and post-processing}.

\subsection{Proposal generation and post-processing}
\label{proposals generation and post-processing}

Now that we have introduced our main contribution, the Voting Evidence Module, we explain how it is used within our proposal generation pipeline to produce $\hat{\Phi}$. 
This requires combining our predicted start/end boundaries to form proposals as well as score and classify these proposals. We describe this next. 

\noindent {\bf Proposal generation.} Given the voting scores $V^{s}$ and $V^{e}$, we consider all start/end times above a predefined threshold, $\xi$, and then consider local maxima as candidate start/end times. We form proposals from every valid start and end combination, i.e. start occurs before end and below a maximum action duration $\tau$. We assign these proposals confidence scores and action classes as follows.

\begin{figure}
	\centering
	\includegraphics[height=110pt]{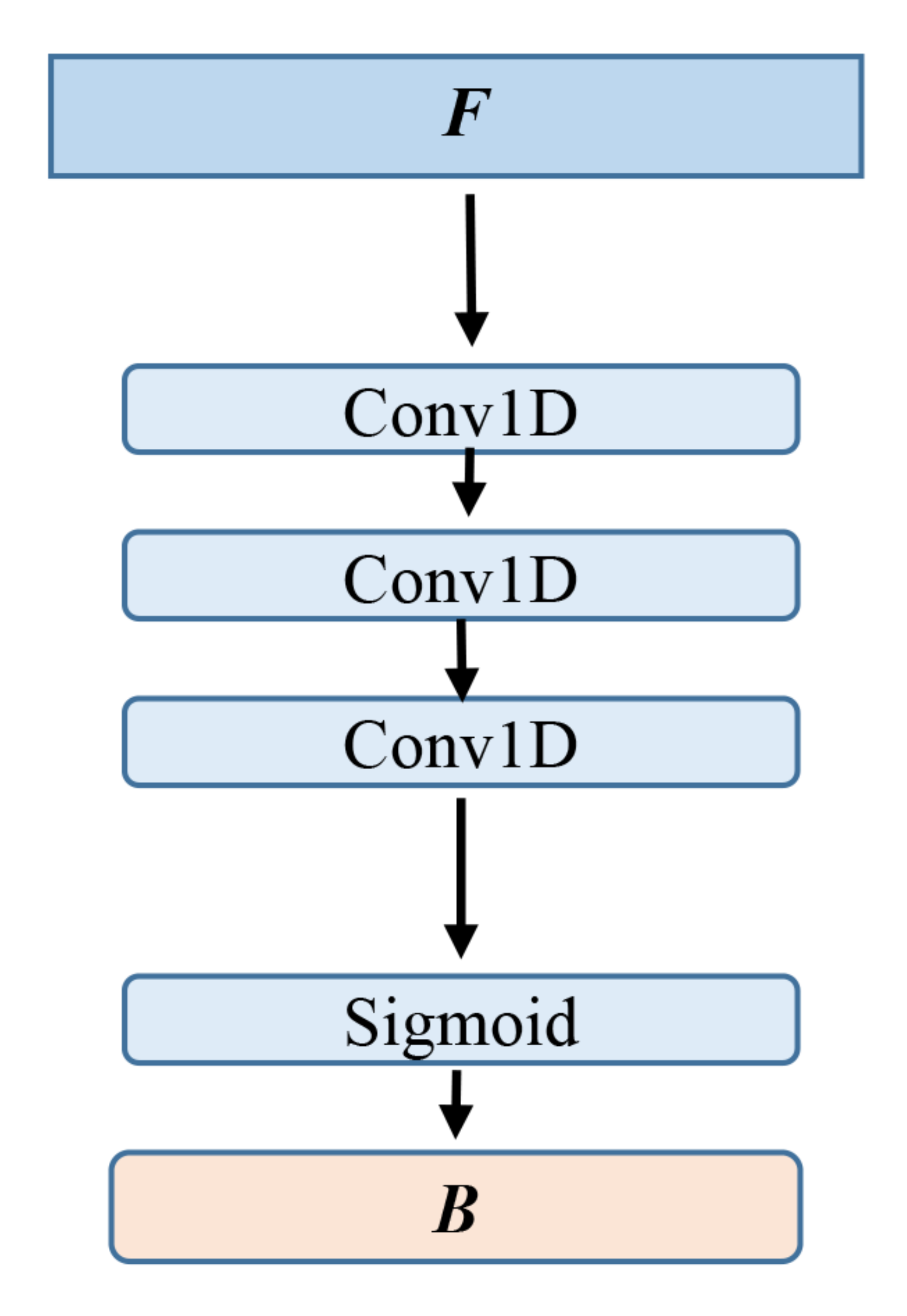} 
	\hspace{20pt}
	\includegraphics[height=110pt]{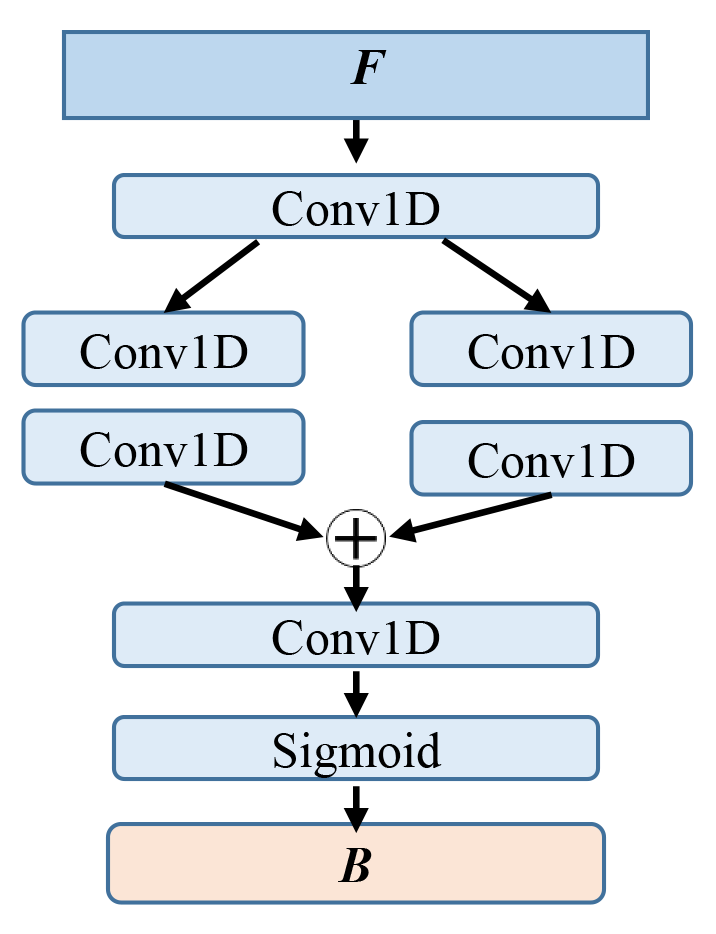}
    \caption{Temporal Evaluation Module from \cite{Lin_2018_ECCV,Lin_2019_ICCV} (left) and our improved version (right). They take in a feature sequence $F$ and output sequences of naive starting ($B^s$) and ending ($B^e$) boundary scores and actionness scores ($B^a$), denoted as $B$.}
	\label{fig:TEM}
\end{figure}

\noindent {\bf Proposal confidence scores.} We begin by processing the feature sequence $F$ with a learned Temporal Evaluation Module (TEM), as used in \cite{Lin_2018_ECCV,Lin_2019_ICCV}. We make slight modifications to the architecture, as shown in Figure \ref{fig:TEM}, by using two 1D-convolutional branches, which performed better experimentally. 
TEM outputs naive boundary starting ($B^{s}$) and ending ($B^{e}$) scores, as well as actionness scores ($B^{a}$). These scores use only local information, attaching the same importance to each frame. 
We use the actionness score $B^{a}$ as background suppression, using element-wise multiplication with the feature sequence. This gives a filtered feature sequence which is the input to VEM. Additionally, we calculate confidence in the proposal directly from its feature sequence $p(\hat s, \hat e)$, using the Proposal Evaluation Module from \cite{Lin_2019_ICCV}.

We fuse $V^s, V^e, B^s, B^e$ and $p(\hat{s}, \hat{e}) $ to calculate confidence scores which are used to rank proposals. The confidence for a single proposal is:
\begin{equation}
\label{voting equation}
\hat c = (v^s_{\hat{s}} + \alpha b^s_{\hat{s}}) (v^e_{\hat{e}} + \alpha b^e_{\hat{e}})
p(\hat s, \hat e)
\end{equation}
where $\alpha$ is a fusion weight, $b^s_{\hat{s}} \in B^s$ is the starting boundary score for $\hat{s}$, and $b^e_{\hat{e}} \in B^e$ is the ending boundary score for $\hat e$ from TEM.

\noindent {\bf Redundant proposal suppression.} Like other action localization works \cite{Lin_2018_ECCV,Lin_2019_ICCV,BSN++}, we use Soft-NMS \cite{softnms} to suppress redundant proposals using the scores $\hat{c}$. 

\noindent {\bf Proposal-to-proposal relations.} We also explore relations between proposals as proposed by PGCN~\cite{PGCN}, through constructing an action proposal graph. We evaluate with/without PGCN, for direct comparison to published works.

\noindent {\bf Classification.} We classify each candidate proposal to obtain the class label $\hat{a}$ and obtain the final prediction set $\hat{\Phi}=\left\{(\hat{s},\ \hat{e},\ \hat{c},\ \hat a) \right\}^M_{m=1}$. 
Note that action classification is performed after proposal generation, making our proposal generation action-independent.

\section{\uppercase{Experiments}}

\subsection{Experimental Setup}
\label{Experimental Setup}

\begin{table*}[t]
\footnotesize
\begin{center}
\caption{Action localization results on ActivityNet-1.3 and THUMOS14. \textbf{Bold} for best model and \underline{underline} for second best.}
\label{main results1}
\begin{tabular}{|ll|cccc|ccccc|}

\hline
\multicolumn{1}{|c}{\multirow {2}{*}{\bf Method}}&\multicolumn{1}{c}{\multirow {2}{*}{\bf Publication}} &\multicolumn{4}{|c}{\bf ActivityNet-1.3 (mAP@IoU)} &\multicolumn{5}{|c|}{\bf THUMOS14 (mAP@IoU)}\\
\cline{3-11}
\multicolumn{1}{|c}{}&\multicolumn{1}{c}{}&\multicolumn{1}{|c}{0.5} & 0.75 & 0.95 & Average &\multicolumn{1}{c}{0.3} & 0.4 & 0.5 & 0.6 & 0.7   \\ 
\hline
SSN \cite{2017SSN} &ICCV2017 &43.26 &28.70 &5.63 &28.28 &51.9 &41.0 &29.8 &- &-  \\[3pt]
TAL-Net \cite{ReFRCNN} &CVPR 2018 &38.23 &18.30 &1.30 &20.22 &53.2 &48.5 &\underline{42.8} &\underline{33.8} &20.8 \\[3pt]

BSN \cite{Lin_2018_ECCV} &ECCV 2018 &46.45 &29.96 &8.02 &30.03 &53.5 &45.0 &36.9 &28.4 &20.0 \\[3pt]
BMN \cite{Lin_2019_ICCV} &ICCV 2019 &50.07 &34.78 &8.29 &33.85 &56.0 &47.4 &38.8 &29.7 &20.5 \\[3pt]
MGG \cite{MGG} &CVPR 2019 &- &- &- &- &53.9 &46.8 &37.4 &29.5 &21.3\\[3pt]
GTAN \cite{2019Gaussian}&CVPR 2019 &$\textbf{52.61}$ &34.14 &8.91 &34.31 &57.8 &47.2 &38.8 &- &- \\[3pt]
G-TAD \cite{GTAD}&CVPR 2020 &50.36 &34.60 &9.02 &34.09 &54.5 &47.6 &40.2 &30.8 &\underline{23.4}\\[3pt]
BC-GNN \cite{BC-GNN} &ECCV 2020 &50.56 &34.75 &\underline{9.37} &34.26 &57.1 &49.1 &40.4 &31.2 &23.1\\[3pt]
BSN++ \cite{BSN++} &AAAI 2021 &51.27 &$\textbf{35.70}$ &8.33 &$\textbf{34.88}$&$\underline{59.9}$ &\underline{49.5} &41.3 &31.9 &22.8\\[3pt]
TVNet &- &\underline{51.35} &\underline{34.96} &$\textbf{10.12}$ &\underline{34.60} &$\textbf{64.7}$ &$\textbf{58.0}$ &$\textbf{49.3}$ &$\textbf{38.2}$ &$\textbf{26.4}$\\[3pt]

\hline
\end{tabular}
\end{center}
\end{table*}

\noindent {\bf Datasets.} We conduct experiments on two temporal action localization datasets: ActivityNet1.3~\cite{Activitynet} and THUMOS14 \cite{THUMOS14} as in \cite{Lin_2018_ECCV,Lin_2019_ICCV,PGCN}. ActivityNet-1.3 consists of 19,994 videos with 200 classes. THUMOS14 contains 413 untrimmed videos with 20 classes for the action localization task.

\noindent {\bf Comparative analysis.} We compare our work to all seminal efforts that evaluate on these two standard benchmarks~\cite{2017SSN,ReFRCNN,Lin_2018_ECCV,Lin_2019_ICCV,MGG,2019Gaussian,PGCN,GTAD,BC-GNN,Uty,BSN++}. As in previous efforts~\cite{PGCN,Uty,GTAD}, we perform an additional test when combining our work with the additional power of proposal-to-proposal relations from PGCN \cite{PGCN} and the temporal aggregation from MUSES \cite{MUSES}.

\noindent {\bf Evaluation Metrics.} We use mean Average Precision (mAP) to evaluate the performance of action localization. To compare to other works, we report the same IoU thresholds. On ActivityNet-1.3 these are \{0.5, 0.75, 0.95\}, and on THUMOS14 they are \{0.3, .., 0.7\}, as well as the average mAP of the IoU thresholds from 0.5 to 0.95 at step size of 0.05.

\noindent {\bf Implementation Details.} For feature extraction, we adopt the two-stream structure \cite{Two-stream} as the visual encoder following previous works \cite{Lin_2018_ECCV,Paul_2018_ECCV,Lin_2019_ICCV,BSN++}.
We parse the videos every 16 frames to extract features as in \cite{Lin_2018_ECCV,Paul_2018_ECCV,Lin_2019_ICCV}.
To unify the various video lengths as input, following previous work \cite{Lin_2018_ECCV,Paul_2018_ECCV}, we sample to obtain a fixed input length, which is $T=100$ for ActivityNet-1.3 and $T=750$ for THUMOS14. The sliding window length is set to $J=15+5$ for ActivityNet-1.3 and $J=10+5$ for THUMOS14.
We first train the TEM, then the PEM following the process in \cite{Lin_2018_ECCV}.
We then train the Voting Evidence Module using the Adam optimizer \cite{Adam} with a learning rate 0.001 for the first 10 epochs and 0.0001 for the remaining 5 epochs for THUMOS14, and 0.0001 and 0.00001 for ActivityNet-1.3. The batch size is set to 512 and 256 for ActivityNet-1.3 and THUMOS14.

For proposal generation, we set the threshold $\xi=0.3$, the maximum action length $\tau=100$ for ActivityNet-1.3, and $\tau=70$ for THUMOS14. 
To ensure a fair comparison for classifying our proposals, we use the same classifier as previous works \cite{Lin_2018_ECCV,Lin_2019_ICCV,GTAD}. We use the top video classification from \cite{2016CUHK} for ActivityNet-1.3. On THUMOS14, we assign the top-2 video classes predicted by \cite{untrimmednets} to all the proposals in that video. For Soft-NMS, we select the top 200 and 400 proposals for ActivityNet-1.3 and THUMOS14, respectively.

\subsection{Results}

\noindent {\bf Main Results.} We compare TVNet with the state-of-the-art methods in Table \ref{main results1}. TVNet achieves comparable performance on ActivityNet-1.3, with a mAP of 10.12\% at an IoU of 0.95, which outperforms all previous methods and shows that our temporal voting can distinguish the boundaries more precisely. On THUMOS14, we exceed all other methods across the range of IoUs commonly reported, for example reaching 49.3\% mAP at 0.5.

Similar to other works, such as \cite{GTAD,Uty}, in Table \ref{main results2} we provide results on THUMOS14 for the PGCN \cite{PGCN} proposal evaluation scheme applied to our proposals. We also provide results combining our proposals with the recently-introduced MUSES evaluation scheme \cite{MUSES}. We report strong performance outperforming all prior works.

\begin{table*}[ht!]
\footnotesize
\begin{center}
\caption{Action localization results on THUMOS14 for methods combined with proposal-to-proposal relations from PGCN \cite{PGCN} and MUSES \cite{MUSES}. \textbf{Bold} for best and \underline{underline} for second best.}
\label{main results2}
\begin{tabular}{|ll|ccccc|}
\hline
\multicolumn{1}{|c}{\multirow {2}{*}{\bf Method}}&\multicolumn{1}{c}{\multirow {2}{*}{\bf Publication}} &\multicolumn{5}{|c|}{\bf THUMOS14 (mAP@IoU)}\\
\cline{3-7}
\multicolumn{1}{|c}{}&\multicolumn{1}{c}{}&\multicolumn{1}{|c}{0.3} & 0.4 & 0.5 & 0.6 & 0.7   \\ 
\hline
BSN + PGCN \cite{PGCN} &ICCV 2019 &63.6 &57.8 &49.1 &- &- \\[3pt]
Uty + PGCN \cite{Uty}  &BMVC 2020 &66.3 &59.8 &50.4  &37.5 &\underline{23.5}\\[3pt]
G-TAD + PGCN\cite{GTAD}&CVPR 2020 &\underline{66.4} &\underline{60.4} &\underline{51.6} &\underline{37.6} &22.9\\[3pt]
TVNet + PGCN &- &$\textbf{68.3}$ &$\textbf{63.7}$ &$\textbf{56.0}$ &$\textbf{39.9}$ &$\textbf{24.2}$\\[3pt]
\hline
\rowcolor{gray!10}{BSN + MUSES \cite{MUSES}} &CVPR 2021 &$\underline{68.9}$ &$\underline{64.0}$ &$\underline{56.9}$ &$\underline{46.3}$ &$\underline{31.0}$ \\[3pt]
\rowcolor{gray!10}TVNet + MUSES \cite{MUSES} &- &$\textbf{71.1}$ &$\textbf{66.4}$ &$\textbf{59.1}$ &$\textbf{47.8}$ &$\textbf{32.1}$  \\[3pt]
\hline
\end{tabular}
\end{center}
\end{table*}
\begin{figure*}[ht!]
	\centering
	\includegraphics[width=0.74\linewidth]{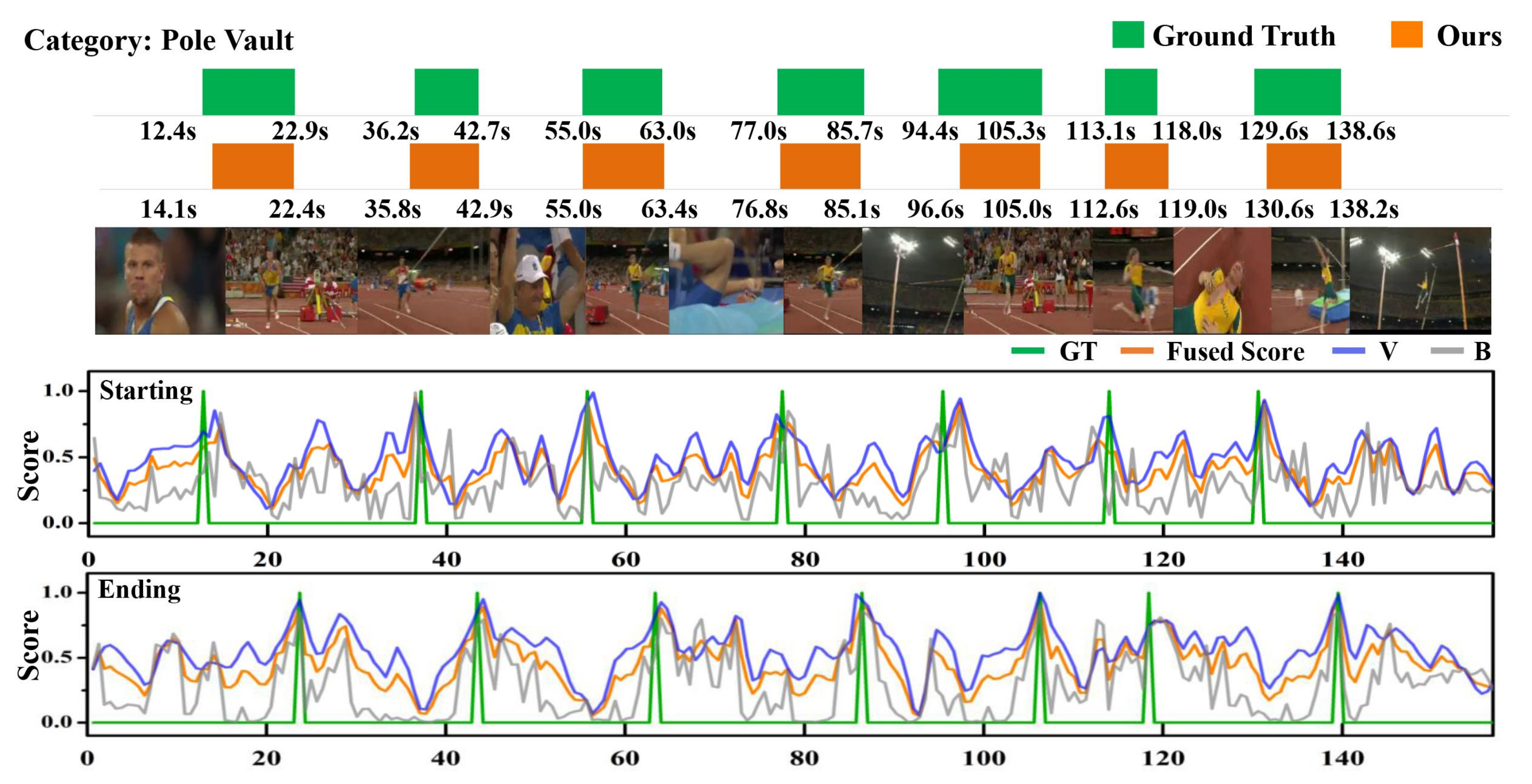} \\
	\vspace*{0.3cm}
	\includegraphics[width=0.74\linewidth]{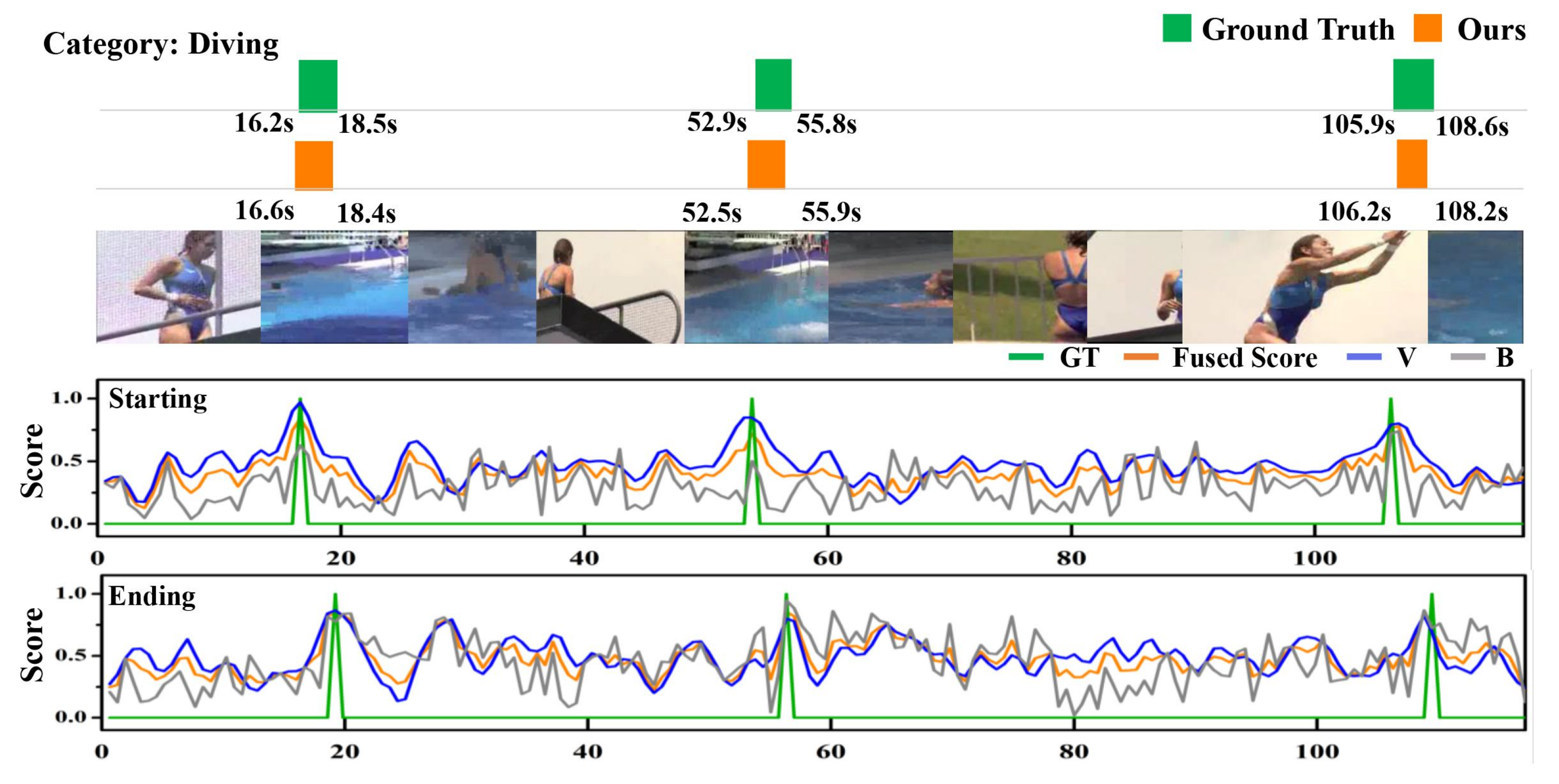} \\
    \caption{Qualitative results on THUMOS14, where TVNet detects multiple dense (top) and sparse (bottom) actions with accurate boundaries. The green bars indicate ground truth instances and the orange bars indicate TVNet detections. The green, orange, blue and grey lines are ground truth boundaries, weighted boundaries scores, voting scores and boundary scores respectively.}
	\label{fig:qualitative-thumos}
\end{figure*}

\begin{figure*}[ht]
	\centering
	\includegraphics[width=0.45\linewidth]{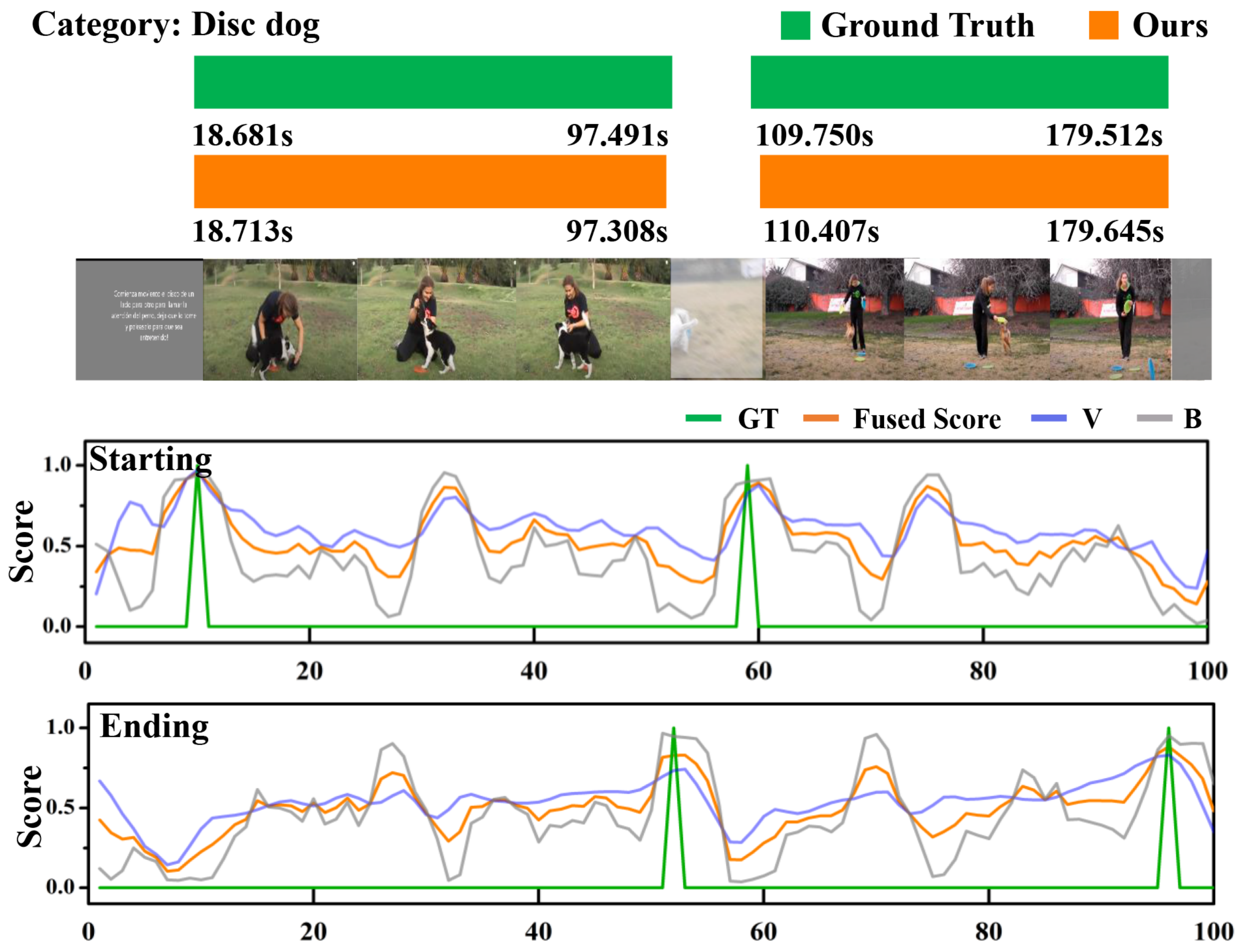}
	\hspace{0cm}
	\includegraphics[width=0.45\linewidth]{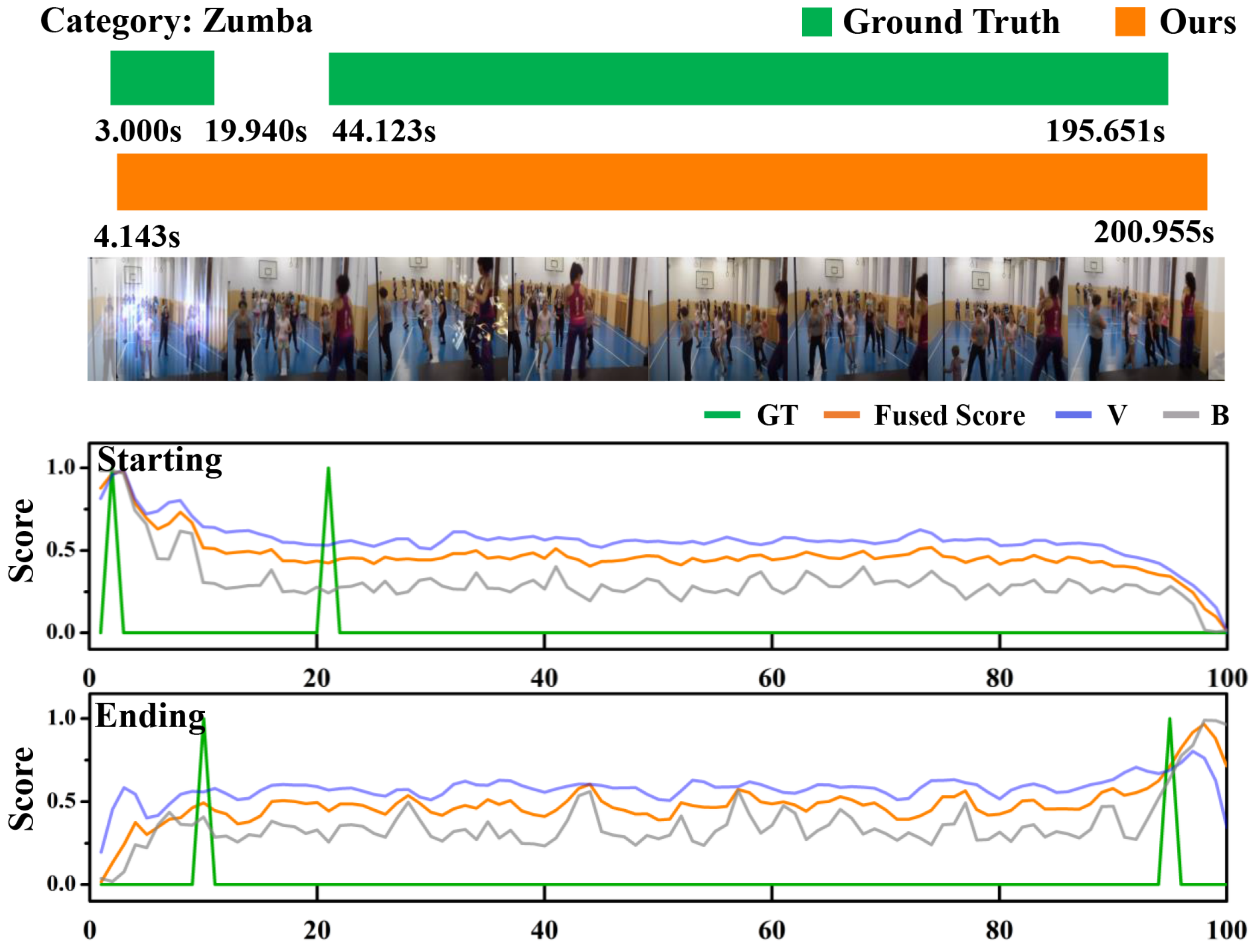}
    \caption{Qualitative TVNet results on ActivityNet.  Left: a success case, where the actions are detected with boundaries closely matching the ground truth. Right: a failure case, where the wrong start/end times are matched forming one long action.}
	\label{fig:qualitative-an}
\end{figure*}

\noindent {\bf Qualitative Results.} We also present qualitative results of TVNet on on both datasets. 
Figure \ref{fig:qualitative-thumos} shows challenging examples on long THUMOS14 videos contain dense (top) and sparse (bottom) short actions. 
Figure \ref{fig:qualitative-an} shows two examples from ActivityNet-1.3. A success case is shown, where long actions are detected with accurate boundaries. A failure case is also shown, where two actions are mistaken as one, but the overall start and end times are accurate.
In all examples, note how the orange boundary curves do not look identical to the peaks in the ground truth. This is desired behaviour, as they are designed to be probabilities found by voting, and their maxima are used to form action proposals which are then evaluated for fused confidence scores.
\section{\uppercase{Ablation Study}}
\label{Ablation Study}
To investigate the behaviour of our model, we
conduct several ablation studies 
\begin{table}[ht!]
\footnotesize
\begin{center}
\caption{The effect of actionness scores ($B^a$) and boundary scores ($B^s, B^e$) on ActivityNet-1.3. The top row indicates neither, which equates to the model with the TEM removed.}
\label{AAM and BAS}
\begin{tabular}{|cc|cccc|}
\hline
\multicolumn{1}{|c}{\multirow {2}{*}{\bf{$B^a$}}}&\multicolumn{1}{c}{\multirow {2}{*}{\bf{$B^s/B^e$}}}
&\multicolumn{4}{|c|}{\bf{mAP@IoU}}\\
\cline{3-6}
\multicolumn{1}{|c}{}&\multicolumn{1}{c}{}&
\multicolumn{1}{|c}{0.5} & 0.75 & 0.95 & Average   \\ 
\hline
 \XSolidBrush &\XSolidBrush &50.45 &34.26 &7.97 &33.55 \\ 
 \Checkmark  &\XSolidBrush &51.30 &34.89 &8.90 &34.40 \\			
 \XSolidBrush &\Checkmark &51.14 &34.74 &9.68 &34.35 \\	
 \Checkmark &\Checkmark &$\textbf{51.35}$ &$\textbf{34.96}$ &$\textbf{10.12}$ &$\textbf{34.60}$ \\
\hline
\end{tabular}
\end{center}
\end{table}
 on ActivityNet-1.3 and THUMOS14 (as in \cite{Lin_2019_ICCV,BSN++}).

\noindent {\bf Effectiveness of TEM.} We test two contributions of the TEM. These are the actionness score $B^a$ and the boundary scores $B^s$ and $B^e$.
Table \ref{AAM and BAS} shows small improvements when using either, and best results when using both. All parts demonstrate an improvement over using neither, which can be considered as the full model without the TEM.

\begin{figure} [ht!]
\centering 
\centering
\includegraphics[height=0.24\textheight]{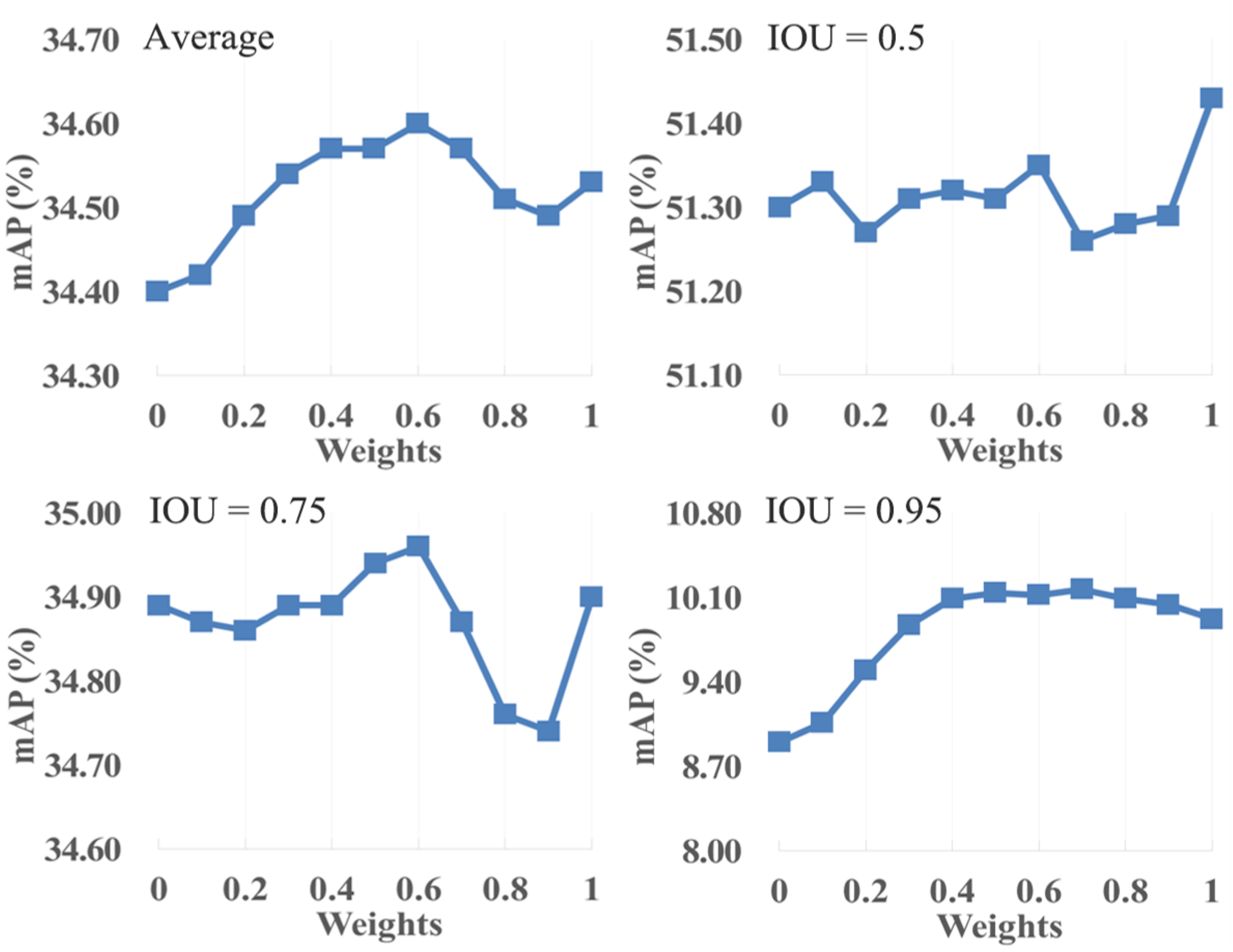}
\caption{Performance of different weights used to fusion on ActivityNet1.3.}
\label{fig:weights}  
\end{figure}

\noindent {\bf Voting and boundary scores.} We fuse the voting scores and the boundary scores to calculate the final boundaries score for each proposal, which is used for ranking. Table \ref{final boundary scores} evaluates the importance of these. Just using the voting score outperforms the boundary score, but the combination of the two is best, suggesting they learn complementary information.
Figure~\ref{fig:weights} shows how performance can be improved by combining these with a suitable weighting. We tried different values from 0 to 1 at step size 0.1 for $\alpha$ in Equation \ref{voting equation}, the best is $\alpha=0.6$.

\begin{table}[ht!]
\footnotesize
\begin{center}
\caption{The effect of different combinations of boundary score ($B$), voting scores ($V$) and proposal generation ($G$) based on $V$ on ActivityNet-1.3. All results are from our implementation, apart from *,  which denotes the original $B$ from \cite{Lin_2019_ICCV}, and can be considered as TVNet without the VEM.}
\label{final boundary scores}
{\begin{tabular}{|ccc|cccc|}
\hline
\multicolumn{1}{|c}{\multirow {2}{*}{\tabincell{c}{\bf{$B$}}}}&\multicolumn{1}{c}{\multirow {2}{*}{\tabincell{c}{\bf{$G$}}}}&\multicolumn{1}{c}{\multirow{2}{*}{\tabincell{c}{\bf{$V$}}}}
&\multicolumn{4}{|c|}{\bf{mAP@IoU}}\\
\cline{4-7}
\multicolumn{1}{|c}{}&\multicolumn{1}{c}{}&\multicolumn{1}{c}{}&
\multicolumn{1}{|c}{0.5} & 0.75 & 0.95 & Average   \\ 
\hline
 \Checkmark* &\XSolidBrush &\XSolidBrush &50.13 &33.18 &9.50 &33.15 \\
 \Checkmark &\XSolidBrush &\XSolidBrush &50.64 &34.30 &8.93 &33.84 \\
 \Checkmark &\Checkmark &\XSolidBrush  &50.68 &34.17 &9.73 &33.87 \\
 \XSolidBrush &\Checkmark &\Checkmark &51.30 &34.89 &8.90 &34.40 \\	
 \Checkmark &\Checkmark &\Checkmark &$\textbf{51.35}$ &$\textbf{34.96}$ &$\textbf{10.12}$ &$\textbf{34.60}$ \\	
\hline
\end{tabular}}
\end{center}
\end{table}

\begin{table*}[ht!]
\footnotesize
\begin{minipage}[t]{0.47\textwidth}
\begin{center}
\caption{The effect of different sliding window lengths $J$ on ActivityNet-1.3.}
\label{window length ANet}
\begin{tabular}{|c|cccc|}
\hline
\multirow{2}{*}{\tabincell{c}{\bf{$J$}}}  & \multicolumn{4}{c|}{\bf{mAP@IoU}}\\ 
\cline{2-5}
& 0.5 & 0.75 & 0.95 & Average   \\ 
\hline
5 &50.84 &34.57 &9.47 &34.08 \\
10 &50.97 &34.61 &9.84 &34.24 \\
15 &$\textbf{51.11}$ &$\textbf{34.63}$ &10.00 &$\textbf{34.36}$ \\
20 &50.97 &34.57 &$\textbf{10.03}$ &34.29 \\
\hline
15 + 5 &$\textbf{51.35}$  &$\textbf{34.96}$  &10.12  &$\textbf{34.60}$ \\
15 + 10 &51.27 &34.87 &9.94 &34.52\\
15 + 20 &51.01 &34.65 &$\textbf{10.13}$ &34.42\\
\hline
\end{tabular}
\end{center}
\end{minipage}
\hspace{0.5cm}
\begin{minipage}[t]{0.47\textwidth}
\begin{center}
\caption{The effect of different sliding window lengths $J$ on THUMOS14.}
\label{window length THUMOS}
\begin{tabular}{|c|ccccc|}
\hline
\multirow{2}{*}{\tabincell{c}{\bf{$J$}}}  & \multicolumn{5}{c|}{\bf{mAP@IoU}}\\ 
\cline{2-6}
& 0.3 & 0.4 & 0.5 & 0.6 & 0.7   \\ 
\hline
5 &$\textbf{64.6}$ &$\textbf{57.9}$ &$\textbf{48.6}$ &$\textbf{38.0}$ &$\textbf{25.6}$ \\
10 &63.9 &56.4 &47.9 &36.2 &25.5 \\
15 &63.6 &56.3 &47.3 &35.4 &24.5 \\
20 &61.9 &54.8 &46.2 &34.8 &24.0 \\
\hline
5 + 10 &$\textbf{64.7}$ &$\textbf{58.0}$ &$\textbf{49.3}$ &$\textbf{38.2}$ &$\textbf{26.4}$\\
5 + 15  &63.8 &56.9 &48.5 &37.3 &25.7 \\
5 + 20 &64.0 &57.0 &47.8 &36.3 &25.0 \\
\hline
\end{tabular}
\end{center}
\end{minipage}
\end{table*}

\begin{table*}[ht!]
\footnotesize
\begin{minipage}[t]{0.47\textwidth}
\begin{center}
\caption{The effect of different maximum action lengths $\tau$ on ActivityNet-1.3.}
\label{MAL ANet}
\begin{tabular}{|c|cccc|}
\hline
\multirow{2}{*}{\tabincell{c}{\bf{$\tau$}}}  & \multicolumn{4}{c|}{\bf{mAP@IoU}}\\ 
\cline{2-5}
& 0.5 & 0.75 & 0.95 & Average   \\ 
\hline
60 &35.88 &9.60 &0.76 &14.44 \\
70 &42.92 &14.29 &1.02 &20.21 \\
80 &47.55 &27.67 &1.70 &26.62 \\
90 &50.42 &33.50 &3.12 &32.10 \\
100 &$\textbf{51.35}$  &$\textbf{34.96}$  &$\textbf{10.12}$  &$\textbf{34.60}$ \\
\hline
\end{tabular}
\end{center}
\end{minipage}
\hspace{0.5cm}
\begin{minipage}[t]{0.47\textwidth}
\begin{center}
\caption{The effect of different maximum action lengths $\tau$ on THUMOS14.}
\label{MAL THUMOS}
\begin{tabular}{|c|ccccc|}
\hline
\multirow{2}{*}{\tabincell{c}{\bf{$\tau$}}}  & \multicolumn{5}{c|}{\bf{mAP@IoU}}\\ 
\cline{2-6}
& 0.3 & 0.4 & 0.5 & 0.6 & 0.7   \\ 
\hline
40 &53.2 &44.2 &32.9 &22.0 &13.7\\
50 &57.8 &50.4 &40.5 &29.0 &18.8 \\
60 &60.7 &53.6 &42.8 &33.5 &23.7\\
70 &$\textbf{64.7}$ &$\textbf{58.0}$ &$\textbf{49.3}$ &$\textbf{38.2}$ &$\textbf{26.4}$\\
80  &64.3 &57.7 &48.3 &37.1 &25.9 \\
\hline
\end{tabular}
\end{center}
\end{minipage}
\end{table*}

\begin{table*}[ht!]
\footnotesize
\begin{minipage}[t]{0.47\textwidth}
\begin{center}
\caption{The effect of different score threshold $\xi$ on ActivityNet-1.3}
\label{Ablation score threshold1}
\begin{tabular}{|c|cccc|}
\hline
\multirow{2}{*}{\tabincell{c}{\bf{$\xi$}}}  & \multicolumn{4}{c|}{\bf{mAP@IoU}}\\ 
\cline{2-5}
& 0.5 & 0.75 & 0.95 & Average   \\ 
\hline
0.1   &51.34 &34.94 &10.09 &$\textbf{34.60}$\\
0.3   &$\textbf{51.35}$  &$\textbf{34.96}$  &$\textbf{10.12}$  &$\textbf{34.60}$\\
0.5   &51.08 &34.73 &9.49 &34.27\\
0.7  &50.72 &34.27 &9.54 &33.96\\
\hline
\end{tabular}
\end{center}
\end{minipage}
\hspace{0.5cm}
\begin{minipage}[t]{0.47\textwidth}
\begin{center}
\caption{The effect of different score threshold $\xi$ on THUMOS14.}
\label{Ablation score threshold2}
\begin{tabular}{|c|ccccc|}
\hline
\multirow{2}{*}{\tabincell{c}{\bf{$\xi$}}}  & \multicolumn{5}{c|}{\bf{mAP@IoU}}\\ 
\cline{2-6}
& 0.3 & 0.4 & 0.5 & 0.6 & 0.7   \\ 
\hline
0.1   &$\textbf{64.8}$ &57.6 &48.4 &36.9 &25.5  \\
0.3   &64.7 &$\textbf{58.0}$ &$\textbf{49.3}$ &$\textbf{38.2}$ &$\textbf{26.4}$\\
0.5  &64.5 &57.4 &48.5 &37.0 &25.6 \\
0.7  &63.7 &56.9 &47.7 &36.6 &25.9 \\
\hline
\end{tabular}
\end{center}
\end{minipage}
\end{table*}

\begin{table}[ht!]
\footnotesize
\begin{center}
\centering
\caption{Ablation studies of the impact of LSTM on ActivityNet-1.3. SRF: Small Receptive Field. SLL: Single Linear Layer.}
\label{Ablation LSTM}
\begin{tabular}{|c|cccc|}
\hline
\multirow{2}{*}{\tabincell{c}{\bf{Layer}}}  & \multicolumn{4}{c|}{\bf{mAP@IoU}}\\ 
\cline{2-5}
& 0.5 & 0.75 & 0.95 & Average   \\ 
\hline
SRF &47.59 &31.20 &3.32 &29.93 \\
SLL &50.29 &34.38 &9.56 &33.97 \\
LSTM &$\textbf{51.35}$  &$\textbf{34.96}$  &$\textbf{10.12}$  &$\textbf{34.60}$ \\
\hline
\end{tabular}
\end{center}
\end{table}

\noindent {\bf Sliding window length.} A larger sliding window length in the VEM could lead to more context information being used for voting at the expense of some missed boundary locations, especially for small action instances. Table \ref{window length ANet} and Table \ref{window length THUMOS} shows that TVNet results are actually relatively stable with respect to sliding window lengths between 5 and 20, with best results at~15+5 for ActivityNet-1.3 and 5+10 for THUMOS14.

\noindent {\bf Maximum  action  length.} This hyperparameter determines the maximum length of candidate proposals. Table \ref{MAL ANet} shows results on ActivityNet-1.3, where all video sequences are scaled to have length less than 100, following standard practice \cite{Lin_2018_ECCV,Lin_2019_ICCV,GTAD,BSN++}. Table \ref{MAL THUMOS} shows results on THUMOS14, where the average video sequence is longer and most action instances are short. The best maximum action length on this dataset ($\tau=70$) is very similar to others ($\tau=64$ for \cite{Lin_2019_ICCV,GTAD,BSN++}). We stop at $\tau=80$ due to GPU memory constraints.

\noindent {\bf Score threshold.} Table \ref{Ablation score threshold1} and Table \ref{Ablation score threshold2} show the effect of varying the threshold $\xi$, which is used to reject potential start and end points if they have low confidence before forming proposals. We use 0.3 for the main experiments, but results are stable for $0.1 < \xi < 0.5$.

\noindent {\bf Effectiveness of LSTM.} To demonstrate the effect of the LSTM when accumulating evidence, Table \ref{Ablation LSTM} shows results when the LSTM is replaced with different architectures. First is a method with a receptive field of 1 (i.e. no temporal accumulation). Second is a single linear layer with the same receptive field as the LSTM. Finally, results from the full LSTM are shown, which performs best across all IoUs. Note that the small receptive field performs very badly on the high IoU.
\section{\uppercase{Conclusion}}
This paper introduced a Temporal Voting Network (TVNet) for temporal action localization. TVNet incorporates a novel Voting Evidence Module, which allows each frame to contribute to boundary localization through voting, whether or not it is a boundary itself.
On ActivityNet-1.3, TVNet achieves significantly better performance than other state-of-the-art methods at IoU of 0.95, highlighting its ability to accurately locate starting and ending boundaries. On THUMOS14, TVNet outperforms other methods when combining its proposals with the powerful PGCN proposal-to-proposal relations and MUSES. We provide qualitative examples, as well as a detailed ablation, which showcases the benefits of our voting-based approach.

\noindent {\bf Acknowledgements:} We use publicly-available datasets. Hanyuan Wang is funded by the China Scholarship Council. Toby Perrett is funded by the SPHERE Next Steps EPSRC Project EP/R005273/1.

\bibliographystyle{apalike}
{\small
\bibliography{example}}

\end{document}